\documentclass[11pt,a4paper]{article}
\usepackage[hyperref]{acl2019}
\usepackage{times}
\usepackage{latexsym}
\usepackage{amsmath}
\usepackage{url}
\usepackage{graphicx}
\usepackage{multirow}
\usepackage{booktabs}
\usepackage{amsfonts}
\usepackage[linguistics]{forest}
\usepackage{verbatim}
\usepackage{hyperref}
\hypersetup{
	urlcolor=blue,
}

\usepackage{tensor}
\aclfinalcopy 

\newcommand{\mat}{impact matrix}

\DeclareMathOperator*{\argmax}{arg\,max}

\title{Perturbed Masking: Parameter-free Probing for Analyzing and Interpreting BERT}

\author{
 Zhiyong Wu$^1$, Yun Chen$^2$, Ben Kao$^1$, Qun Liu$^3$\\
 $^1$The University of Hong Kong, Hong Kong, China \\
 $^2$Shanghai University of Finance and Economics, Shanghai, China\\
 $^3$Huawei Noah's Ark Lab, Hong Kong, China \\
 \{zywu,kao\}@cs.hku.hk, yunchen@sufe.edu.cn, qun.liu@huawei.com\\
}

\date{}

\begin{document}
\maketitle
\begin{abstract}
By introducing a small set of additional parameters, a {\it probe} learns to solve specific linguistic tasks (e.g., dependency parsing) in a supervised manner using feature representations (e.g., contextualized embeddings). The effectiveness of such {\it probing tasks} is taken as evidence that the pre-trained model encodes linguistic knowledge. However, this approach of evaluating a language model is undermined by the uncertainty of the amount of knowledge that is  learned by the probe itself. 
Complementary to those works, we propose a parameter-free probing technique for analyzing pre-trained language models (e.g., BERT). 
Our method does not require direct supervision from the probing tasks, nor do we introduce additional parameters to the probing process. Our experiments on BERT show that syntactic trees recovered from BERT using our method are significantly better than linguistically-uninformed baselines. We further feed the empirically induced dependency structures into a downstream sentiment classification task and find its improvement compatible with or even superior to a human-designed dependency schema.\footnote{\url{https://github.com/LividWo/Perturbed-Masking}}

\end{abstract}

\section{Introduction}

Recent prevalent pre-trained language models such as ELMo~\citep{Peters2018DeepCW}, BERT~\citep{Devlin2018BERTPO}, and XLNet~\citep{Yang2019XLNetGA} achieve state-of-the-art performance for a diverse array of downstream NLP tasks. 
An interesting area of research is to investigate the interpretability of these pre-trained models (i.e., the linguistic properties they capture).
Most recent approaches are built upon the idea of {\it probing classifiers}~\cite{Shi2016, Adi2017, conneau2018you, peters2018dissecting, Hewitt2019, Clark2019, Tenney2019, Jawahar2019}.
A {\it probe} is a simple neural network (with a small additional set of parameters) that uses the feature representations generated by a pre-trained model (e.g., hidden state activations, attention weights) and is trained to perform a supervised task (e.g., dependency labeling). The performance of a {\it probe} is used to measure the quality of the generated representations with the assumption that the measured quality is mostly attributable to the pre-trained language model.

One downside of such approach, as pointed out in~\cite{hewitt2019designing}, is that a probe introduces a new set of additional parameters, which makes the results difficult to interpret. Is it the pre-trained model that captures the linguistic information, or is it the probe that learns the downstream task itself and thus encodes the information in its additional parameter space?

In this paper we propose a parameter-free probing technique called {\it Perturbed Masking} to analyze and interpret pre-trained models. The main idea is to introduce the {\it Perturbed Masking} technique into the masked language modeling (\textbf{MLM}) objective to measure the impact a word $x_j$ has on predicting another word $x_i$ (Sec~\ref{sec:token-mask}) and then induce the global linguistic properties (e.g., dependency trees) from this inter-word information.

Our contributions are threefold:

\noindent $\bullet$ We introduce a new parameter-free probing technique, \textit{Perturbed Masking}, to estimate inter-word correlations. Our technique enables global syntactic information extraction. 

\noindent $\bullet$  We evaluate the effectiveness of our probe over a number of linguistic driven tasks (e.g., syntactic parsing, discourse dependency parsing). Our results reinforce the claims of recent probing works, and further complement them by quantitatively evaluating the validity of their claims.

\noindent $\bullet$ We feed the empirically induced dependency structures into a downstream task to make a comparison with a parser-provided, linguist-designed dependency schema and find that our structures perform on-par or even better (Sec~\ref{sec:sentiment}) than the parser created one. This offers an insight into the remarkable success of BERT on downstream tasks.

\section{Perturbed Masking}

We propose the perturbed masking technique to assess the impact one word has on the prediction of another in MLM. The  inter-word information derived serves as the basis for our later analysis. 

\subsection{Background: BERT}
BERT\footnote{In our experiments, we use the base, uncased version from \cite{Wolf2019HuggingFacesTS}. }~\cite{Devlin2018BERTPO} is a large Transformer network that is pre-trained on 3.3 billion tokens of English text. It performs two tasks: (1) Masked Language Modeling  (\textbf{MLM}): randomly select and mask 15\% of all tokens in each given sequence, and then predict those masked tokens. In masking, a token is (a) replaced by the special token [MASK], (b) replaced by a random token, or (c) kept unchanged. These replacements are chosen 80\%, 10\%, and 10\% of the time, respectively. (2)Next Sentence Prediction: given a pair of sentences, predict whether the second sentence follows the first in an original document or is taken from another random document.

\subsection{Token Perturbation}
\label{sec:token-mask}

Given a sentence as a list of tokens $\mathbf{x}=[x_1, \ldots, x_T]$, BERT maps each $x_i$ into a contextualized representation $H_{\theta}(\mathbf{x})_i$, where $\theta$ represents the network's parameters. Our goal is to derive a function $f(x_i, x_j)$ that captures the impact a context word $x_j$ has on the prediction of another word $x_i$. 

We propose a two-stage approach to achieve our goal. 
First, we replace $x_i$ with the [MASK] token and feed the new sequence $\mathbf{x} \backslash \{x_i\} $ into BERT. We use $H_{\theta}(\mathbf{x} \backslash \{x_i\})_i$ to denote the representation of $x_i$. To calculate the impact $x_j \in \mathbf{x} \backslash \{x_i\} $ has on $H_{\theta}(\mathbf{x} \backslash \{x_i\})_i$, we further mask out $x_j$ to obtain the second corrupted sequence $\mathbf{x} \backslash \{x_i, x_j\} $. Similarly, $H_{\theta}(\mathbf{x} \backslash \{x_i, x_j\})_i$ denotes the new representation of token $x_i$. 

We define $f(x_i, x_j)$ as:
$$ f(x_i, x_j) = d \left(H_{\theta}(\mathbf{x} \backslash \{x_i\})_i,  H_{\theta}(\mathbf{x} \backslash \{x_i, x_j\})_i\right)
$$
where $d(\mathbf{x}, \mathbf{y})$ is the distance metric that captures the difference between two vectors. We experimented with two options for $d(\mathbf{x}, \mathbf{y})$: 

\noindent $\bullet$ \textbf{Dist:} Euclidean distance between $\mathbf{x}$ and $\mathbf{y}$

\noindent $\bullet$ \textbf{Prob}: $ d(\mathbf{x}, \mathbf{y}) = a(\mathbf{x})_{x_i} - a(\mathbf{y})_{x_i}$,
\newline
where $a(\cdot)$ maps a vector into a probability distribution among the words in the vocabulary. $a(\mathbf{x})_{x_i}$ represents the probability of predicting token $x_i$ base on $\mathbf{x}$.

By repeating the two-stage perturbation on each pair of tokens $x_i, x_j \in \mathbf{x}$ and calculating $f(x_i, x_j)$, we obtain an \textbf{impact matrix} $\mathcal{F}$, where $\mathcal{F}_{ij} \in \mathbb{R}^{T \times T}$. 
Now, we can derive algorithms to extract syntactic trees from $\mathcal{F}$ and compare them with ground-truth trees that are obtained from benchmarks.
Note that BERT uses byte-pair encoding~\cite{sennrich2016subword} and may split a word into multiple tokens(subwords). To evaluate our approach on word-level tasks, we make the following changes to obtain inter-word impact matrices. In each perturbation, we mask all tokens of a split-up word.
The impact \textit{on} a split-up word is obtained by averaging\footnote{We also experimented with other alternatives, but observe no significant difference.} the impacts over the split-up word's tokens. To measure the impact exerted \textit{by} a split-up word, we assume the impacts given by its tokens are the same; We use the impact given by the first token for convenience.

\subsection{Span Perturbation}
\label{sec:clause-mask}
Given the token-level perturbation above, it is straightforward to extend it to span-level perturbation. We investigate how BERT models the relations between spans, which can be phrases, clauses, or paragraphs. As a preliminary study, we investigate how well BERT captures document structures.

We model a document $D$ as N non-overlapping text spans 
$D = [e_1, e_2, \ldots, e_N]$, where each span $e_i$ contains a sequence of tokens $e_i = [x_1^i, x_2^i, \ldots, x_M^i]$. 

For span-level perturbation, instead of masking one token at a time, we mask an array of tokens in a span simultaneously. We obtain the span representation by averaging the representations of all the tokens the span contains. Similarly, we calculate the impact $e_j$ has on $e_i$ by:
$$ f(e_i, e_j) = d \left(H_{\theta}(D \backslash \{e_i\})_i,  H_{\theta}(D \backslash \{e_i, e_j\})_i\right)
$$
where $d$ is the \textbf{Dist} function.

\section{Visualization with Impact Maps}
\label{sec:visualization}
Before we discuss specific syntactic phenomena, let us first analyze some example impact matrices derived from sample sentences. We visualize an impact matrix of a sentence by displaying a heatmap. We use the term ``impact map'' to refer to a heatmap of an impact matrix.

\begin{figure}
    \centering
    \includegraphics[width=0.5\textwidth]{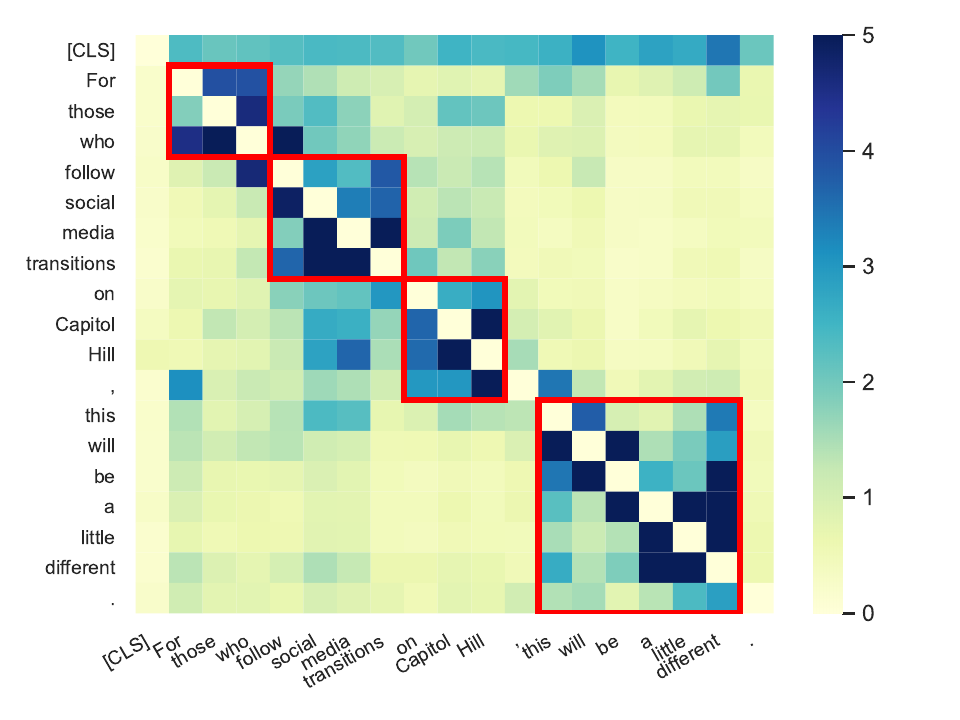}
    \caption{Heatmap of the \mat\ for the sentence ``For those who follow social media transitions on Capitol Hill, this will be a little different.''}
    \label{fig:heatmap}
\end{figure}

\textbf{Setup.} We extract impact matrices by feeding BERT with 1,000 sentences from the English Parallel Universal Dependencies (PUD) treebank of the CoNLL 2017 Shared Task~\cite{zeman2017conll}. We follow the setup and pre-processing steps employed in pre-training BERT. An example impact map is shown in Figure~\ref{fig:heatmap}.

\textbf{Dependency.} We notice that the impact map contains many \textit{stripes}, which are short series of vertical/horizontal cells, typically located along the diagonal. 
Take the word ``\textit{different}'' as an example (which is illustrated by the second-to-last column in the impact matrix). We observe a clear vertical stripe above the main diagonal. The interpretation is that this particular occurrence of the word ``\textit{different}'' strongly affects the occurrences of those words before it. These strong influences are shown by the darker-colored pixels seen in the second last column of the impact map. This observation agrees with the ground-truth dependency tree, which selects ``\textit{different}'' as the head of all remaining words in the phrase ``\textit{this will be a little different}.'' We also observe similar patterns on ``\textit{transitions}'' and ``\textit{Hill}''. Such correlations lead us to explore the idea of extracting dependency trees from the matrices (see Section~\ref{sec:dependency}).

\begin{figure}
    \centering
    \resizebox{0.45\textwidth}{!}{
    \begin{forest}
    [,for tree={s sep=3mm, inner sep=0, l=0.6em}, delay={where content={}{shape=coordinate}{}}
        [follow, tier=word]
        [
            [   [social, tier=word][media, tier=word][transitions, tier=word]    ]
            [   [on, tier=word]
                [   [Capitol, tier=word] [Hill, tier=word]    ]
            ]
        ]
    ]
    \end{forest}
}
    \caption{Part of the constituency tree.}
    \label{fig:goldtree}
\end{figure}
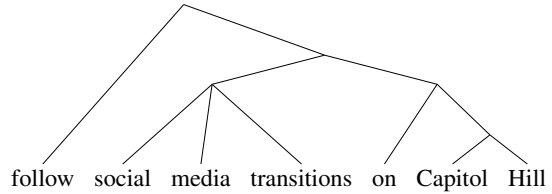

\textbf{Constituency.} Figure~\ref{fig:goldtree} shows part of the constituency tree of our example sentence generated by Stanford CoreNLP~\cite{corenlp}. In this sentence,  ``\textit{media}'' and ``\textit{on}'' are two words that are adjacent to  ``\textit{transitions}''. From the tree, however, we see that ``\textit{media}'' is closer to ``\textit{transitions}'' than ``\textit{on}'' is in terms of syntactic distance. If a model is syntactically uninformed, we would expect ``\textit{media}'' and ``\textit{on}'' to have comparable impacts on the prediction of ``\textit{transitions}'', and vice versa. However, we observe a far greater impact (darker color) between ``\textit{media}'' and ``\textit{transitions}'' than that between ``\textit{on}'' and ``\textit{transitions}''. We will further support this observation with empirical experiments in Section~\ref{sec:constituency}. 

\textbf{Other Structures.} Along the diagonal of the impact map, we see that words are grouped into four contiguous chunks that have specific intents (e.g., a noun phrase -- \textit{on Capitol Hill}). We also observe that the two middle chunks have relatively strong inter-chunk word impacts and thus a bonding that groups them together, forming a larger verb phrase. This observation suggest that BERT may capture the compositionality of the language.

In the following sections we quantitatively evaluate these observations. 
\section{Syntactic Probe} 

We start with two syntactic probes -- dependency probe and constituency probe.

\subsection{Dependency Probe}
\label{sec:dependency}
With the goal of exploring the extent dependency relations are captured in BERT, we set out to answer the following question:  Can BERT outperform linguistically uninformed baselines in unsupervised dependency parsing? If so, to what extent? 

We begin by using the token-level perturbed masking technique to extract an impact matrix $\mathcal{F}$ for each sentence. We then utilize graph-based algorithms to induce a dependency tree from $\mathcal{F}$, and compare it against ground-truth whose annotations are linguistically motivated. 

\paragraph{Experiment Setup.}
We evaluate the induced trees on two benchmarks: (1) the PUD treebank described in Section~\ref{sec:visualization}. (2) the WSJ10 treebank, which contains 7,422 sentences (all less than 10 words after punctuation removal) from the Penn Treebank (PTB)~\cite{marcus1993penn}. Note that the original PTB does not contain dependency annotations. Thus, we convert them into Universal Dependencies using Stanford CoreNLP. We denote this set as WSJ10-U. 

Next, two parsing algorithms, namely, the Eisner algorithm~\shortcite{eisner1996three} and Chu-Liu/Edmonds (CLE) algorithm~\shortcite{chu1965shortest, edmonds1967optimum}, are utilized to extract the projective and non-projective unlabeled dependency trees, respectively. Given that our impact matrices have no knowledge about the dependency root of the sentence, we use the gold root in our analysis. Introducing the gold root may artificially improve our results slightly. We thus apply this bias evenly across all baselines to ensure a fair comparison, as done in~\cite{Tiedemann2018, htut2019attention}. 

We compared our approach against the following baselines: (1) right-(left-) chain baseline, which always selects the next(previous) word as dependency head. (2) A \textit{random} BERT baseline, with which we randomly initialize weights of the BERT model~\cite{htut2019attention}, then use our methods to induce dependency trees.

We measure model performance using Unlabeled Attachment Score (UAS). We note that UAS has been shown to be highly sensitive to annotation variations~\cite{schwartz2011neutralizing, tsarfaty2011ned, kubler2009dependency}. Therefore, it may not be a fair evaluation metric for analyzing and interpreting BERT. To reflect the real quality of the dependency structures that are retained in BERT, we also report Undirected UAS (UUAS)~\cite{klein2004dmv} and the Neutral Edge Direction (NED) scores~\cite{schwartz2011neutralizing}.

\begin{table}[]
    \centering
    \begin{tabular}{lcc}
    \toprule
     \multirow{2}{*}{\textbf{Model}}   & \multicolumn{2}{c}{\textbf{Parsing UAS}}\\
                               & \multicolumn{1}{c}{\textbf{WSJ10-U}} & \multicolumn{1}{c}{\textbf{PUD}} \\
    \toprule
     Right-chain      & 49.5  & 35.0 \\ 
     Left-chain       & 20.6  & 10.7 \\
     Random BERT      & 16.9  & 10.2 \\
     \toprule
     Eisner+Dist      & \textbf{58.6}  & \textbf{41.7}\\
     Eisner+Prob      & 52.7  & 34.1\\
     CLE+Dist         & 51.5  & 33.2\\
    \toprule
    \end{tabular}
    \caption{UAS results of BERT on unsupervised dependency parsing. }
    \label{tab:dep-result}
\end{table}

\textbf{Results.}  Tables~\ref{tab:dep-result} and \ref{tab:new-metric} show the results of our dependency probes. From Table~\ref{tab:dep-result}, we see that although BERT is trained without any explicit supervision from syntactic dependencies, to some extent the syntax-aware representation already exists in it. The best UAS scores it achieves (Eisner+Dist) are substantially higher than that of the random BERT baseline with respect to both WSJ10-U(+41.7) and PUD(+31.5). Moreover, the \textit{Dist} method significantly outperforms the \textit{Prob} method on both datasets we evaluated. We thus use \textit{Dist} as the default distance function in our later discussion. We also note that the Eisner algorithm shows a clear advantage over CLE since English sentences are mostly projective. However, our best performing method does not go much beyond the strong right-chain baseline (with gold root modified), showing that the dependency relations learned are mostly those simple and local ones.  

For reference, the famous unsupervised parser -- DMV~\cite{klein2004dmv} achieves a 43.2 UAS on WSJ10 with Collins~\shortcite{collins1999head} conventions. Note that the DMV parser utilizes POS tags for training while ours start with the gold root. The results are therefore not directly comparable. By putting them together, however, we see potential room for improvement for current neural unsupervised dependency parsing systems in the BERT era.

\begin{table}[]
    \centering
    \begin{tabular}{lccc}
    \toprule
    \textbf{Model} & \textbf{UAS} & \textbf{UUAS} & \textbf{NED} \\
    \toprule 
    Eisner+Dist         & 41.7  & 52.1  & 69.6 \\
    Right-chain         & 35.0  & 39.9  & 41.2 \\ 
    \toprule
    \end{tabular}
    \caption{Performance on PUD when evaluated using UAS, UUAS, and NED.}
    \label{tab:new-metric}
\end{table}

From Table~\ref{tab:new-metric}, we see that although BERT only outperforms the right-chain baseline modestly in terms of UAS, it shows significant improvements on UUAS (+12.2) and NED (+28.4). We also make similar observation with WSJ10-U. 
This suggests that BERT does capture inter-word dependencies despite that it may not totally agree with one specific human-designed governor-dependent schema.
We manually inspect those discrepancies and observe that they can also be syntactically valid. For instance, consider the sentence ``It closed on Sunday.''. For the phrase ``on Sunday'', our method selects the functional word ``on'' as the head while the gold-standard annotation uses a lexical head (``Sunday'')\footnote{This specific choice is actually agreed with the YM \cite{yamada2003statistical} schema.}.

The above findings prove that BERT has learned its own syntax as a by-product of self-supervised training, not by directly copying any human design. However, giving the superior performance of BERT on downstream tasks, it is natural to ask if BERT is learning an empirically useful structure of language. We investigate this question in Sec~\ref{sec:sentiment}.

\subsection{Constituency Probe}
\label{sec:constituency}

We now examine the extent BERT learns about the constituent structure of sentences. We first present the algorithm for unsupervised constituent parsing, which executes in a top-down manner by recursively splitting larger constituents into smaller ones. 

\paragraph{Top-Down Parsing.}
Given a sentence as a sequence of tokens $\mathbf{x}=[x_1, \ldots, x_T]$ and the corresponding \mat\ $\mathcal{F}$. We start by finding the best splitting position $k$ that will separate the sentence into constituents  $( (\mathbf{x}_{<k} ), (x_k, (\mathbf{x}_{>k} ) ))$, where $\mathbf{x}_{<k} = [x_1, \ldots, x_{k-1}]$. The best splitting position ensures that each constituent has a large average impact between words within it (thus those words more likely to form a constituent) while at the same time the impact between words of different constituents are kept as small as possible (thus they are unlikely to be in the same constituent).  Mathematically, we decide the best $k$ for the constituent $\mathbf{x}=[x_i, x_{i+1}, \ldots, x_j]$ by the following optimization:
\begin{equation}
\begin{split}
    \argmax_k & \quad \mathcal{F}_{i,\ldots k}^{i,\ldots ,k}   + \mathcal{F}_{k+1,\ldots  ,j}^{k+1,\ldots  ,j} \\
& - \mathcal{F}_{i, \ldots, k}^{k+1,\ldots ,j} - \mathcal{F}_{k+1,\ldots, j}^{i, \ldots, k}
 \end{split}
\end{equation}

where $\mathcal{F}_{i,\ldots k}^{i,\ldots ,k} = \frac{\sum_{a=i}^{k}\sum_{b=i}^{k} f(x_a,x_b)}{|\theta|} $, and $|\theta|$ is the number of off-diagonal elements in the corresponding impact matrix $
\left[
\begin{matrix}
x_{i, i} & ... & x_{i, k} \\
\vdots & \ddots & \vdots \\
x_{k, i} & ... & x_{k, k}
\end{matrix}
\right] 
$. 
We recursively split $ (\mathbf{x}_{<k} )$ and $(\mathbf{x}_{>k} )$ until only single words remain. Note that this top-down strategy is similar to that of ON-LSTM~\cite{shen2018ordered} and PRPN~\cite{shen2018prpn}, but differs from them in that ON-LSTM and PRPN decide the splitting position based on a ``syntactic distance vector'' which is explicitly modeled by a special network component. To distinguish our approach from the others, we denote our parser as \textbf{MART} (\textbf{MA}t\textbf{R}ix-based \textbf{T}op-down parser)

\begin{table*}[]
    \centering
    \begin{tabular}{lccccccc}
    \toprule
    \multirow{2}{*}{\textbf{Model}} & \multicolumn{2}{c}{\textbf{Parsing F1}}   & \multicolumn{5}{c}{\textbf{Accuracy on PTB23 by Tag}}  \\
                                    & \textbf{WSJ10}    & \textbf{PTB23}    & \textbf{NP} & \textbf{VP} & \textbf{PP} & \textbf{S} & \textbf{SBAR}\\ 
    \toprule
    PRPN-LM                & 70.5  & 37.4  & 63.9     & -  & 24.4  & -     & - \\
    ON-LSTM 1st-layer      & 42.8  & 24.0  & 23.8   & 15.6  & 18.3  & 48.1  & 16.3\\
    ON-LSTM 2nd-layer      & 66.8  & 49.4  & 61.4  & 51.9  & 55.4  & 54.2  & 15.4\\
    ON-LSTM 3rd-layer      & 57.6  & 40.4  & 57.5  & 13.5  & 47.2  & 48.6  & 10.4\\
    300D ST-Gumbel w/o Leaf GRU     & -      & 25.0  & 18.8  & -   & 9.9   & -     & -\\ 
    300D RL-SPINN w/o Leaf GRU      & -     & 13.2  & 24.1  & -  & 14.2  & -     & -\\
    \toprule
    \textbf{MART}           & 58.0  & 42.1  & 44.6  & 47.0  & 50.6  & 66.1  & 51.9 \\
    Right-Branching      & 56.7  & 39.8  & 25.0 & 71.8 & 42.4 & 74.2 & 68.8 \\
    Left-Branching       & 19.6  & 9.0   & 11.3 & 0.8 & 5.0 & 44.1 & 5.5 \\
    \toprule                        
    \end{tabular}
    \caption{Unlabeled parsing F1 results evaluated on WSJ10 and PTB23.}
    \label{tab:constituent}
\end{table*}

\paragraph{Experiment Setup.}
We follow the experiment setting in Shen et al~\shortcite{shen2018ordered, shen2018prpn} and evaluate our method on the 7,422 sentences in WSJ10 dataset and the PTB23 dataset (the traditional PTB test set for constituency parsing).

\paragraph{Results.} Table~\ref{tab:constituent} shows the results of our constituency probes. From the table, we see that BERT outperforms most baselines on PTB23, except for the second layer of ON-LSTM. Note that all these baselines have specifically-designed architectures for the unsupervised parsing task, while BERT's knowledge about constituent formalism emerges purely from self-supervised training on unlabeled text. 

It is also worth noting that recent results~\cite{dyer2019critical, li2019imitation} have suggested that the parsing algorithm used by ON-LSTM (PRPN) is biased towards the right-branching trees of English, leading to inflated F1 compared to unbiased parsers. To ensure a fair comparison with them, we also introduced this right-branching bias. However, our results show that our method is also robust without this bias (e.g., only 0.9 F1 drops on PTB23).

To further understand the strengths and weaknesses of each system, we analyze their accuracies by constituent tags. In Table~\ref{tab:constituent}, we show the accuracies of five most common tags in PTB23. We find that the success of PRPN and ON-LSTM mainly comes from the accurate identification of NP (noun phrase), which accounts for 38.5\% of all constituents. For other phrase-level tags like VP (verb phrase) and PP (prepositional phrase), the accuracies of BERT are competitive. Moreover, for clause level tags, BERT significantly outplays ON-LSTM. Take SBAR (clause introduced by a subordinating conjunction) for example, BERT achieves an accuracy of 51.9\%, which is about 3.4 times higher than that of ON-LSTM. One possible interpretation is that BERT is pre-trained on long contiguous sequences extracted from a document-level corpus. And the masking strategy (randomly mask 15\% tokens) utilized may allow BERT to learn to model a sequence of words (might form a clause).

\section{Discourse Probe}
\label{sec:discourse}
Having shown that clause-level structures are well-captured in BERT using the constituency probe, we now explore a more challenging probe -- probing BERT's knowledge about the structure of a document. 
A document contains a series of coherent text spans, which are named Elementary Discourse Units (EDUs)~\cite{yang2018scidtb, polanyi1988formal}. EDUs are connected to each other by discourse relations to form a document. 
We devise a discourse probe to investigate how well BERT captures structural correlations between EDUs. As the foundation of the probe, we extract an EDU-EDU impact matrix for each document using span-level perturbation. 

\paragraph{Setup.} We evaluate our probe on the discourse dependency corpus SciDTB~\cite{yang2018scidtb}. We do not use the popular discourse corpora RST-DT~\cite{carlson2003builrst} and PDTB~\cite{prasad2008pdtb} because PDTB focuses on local discourse relations but ignores the whole document structure, while RST-DT introduces intermediate nodes and does not cover non-projective structures. We follow the same baseline settings and evaluation procedure in Sec~\ref{sec:dependency}, except that we remove gold root from our evaluation since we want to compare the accuracy by syntactic distances. 

\begin{table}[]
    \centering
    \resizebox{\linewidth}{!}{
    \begin{tabular}{lccccc}
    \toprule
     \multirow{2}{*}{\textbf{Model}} & \multirow{2}{*}{\textbf{UAS}}  & \multicolumn{4}{c}{\textbf{Accuracy by distance}}\\
                    &               & 0 & 1 & 2 & 5 \\  
     \toprule
     Right-chain    & 10.7 & 20.5 & - & - & - \\
     Left-chain     & \textbf{41.5} & \textbf{79.5} & - & - & - \\
     Random BERT    & 6.3 & 20.4 & 7.5 & 3.5 & 0.0 \\ 
     Eisner+Dist    & 34.2 & 61.6 & 7.3 & \textbf{7.6} & \textbf{12.8}\\
     CLE+Dist       & 34.4 & 63.8 & 3.3 & 3.5 & 2.6 \\

    \toprule
    \end{tabular}
    }
    \caption{Performance of different discourse parser. The distance is defined as the number of EDUs between head and dependent.}
    \label{tab:discourse}
\end{table}
\paragraph{Results.} Table~\ref{tab:discourse} shows the performance of our discourse probes. We find that both Eisner and CLE achieve significantly higher UAS (+28) than the random BERT baseline. This suggests that BERT is aware of the structure of the document it is given. In particular, we observe a decent accuracy in identifying discourse relations between adjacent EDUs, perhaps due to the ``next sentence prediction'' task in pre-training, as pointed out in~\cite{shi2019next}.  However, our probes fall behind the left-chain baseline, which benefits from its strong structural prior\footnote{For reference, a supervised graph-based parser~\cite{li2014text} achieves an UAS of 57.6 on SciDTB} (principal clause mostly in front of its subordinate clause). Our finding sheds some lights on BERT's success in downstream tasks that have paragraphs as input (e.g., Question Answering). 

\section{BERT-based Trees VS Parser-provided Trees}
\label{sec:sentiment}

Our probing results suggest that although BERT has captured a certain amount of syntax, there are still substantial disagreements between the syntax BERT learns and those designed by linguists. For instance, our constituency probe on PTB23 significantly outperforms most baselines, but it only roughly agree with the PTB formalism (41.2\% F1). However, BERT has already demonstrated its superiority in many downstream tasks. An interesting question is whether \textit{BERT is learning an empirically useful or even better structure of a language}.

To answer this question, we turn to neural networks that adopt dependency parsing trees as the explicit structure prior to improve downstream tasks. We replace the ground-truth dependency trees those networks used with ones induced from BERT and approximate the effectiveness of different trees by the improvements they introduced.

We conduct experiments on the Aspect Based Sentiment Classification (\textbf{ABSC}) task~\cite{absc}. ABSC is a fine-grained sentiment classification task aiming at identifying the sentiment expressed towards each aspect of a given target entity. As an example, in the following comment of a restaurant, ``I hated their fajitas, but their salads were great'', the sentiment polarities for aspect \textit{fajitas} is negative and that of \textit{salads} is positive. It has been shown in \citet{zhang2019aspect} that injecting syntactic knowledge into neural networks can improve ABSC accuracy. Intuitively, given an aspect, a syntactically closer context word should play a more important role in predicting that aspect's sentiment. They integrate the distances between context words and the aspect on a dependency tree into a convolution network and build a Proximity-Weighted Convolution Network (PWCN). As a naive baseline, they compare with network weighted by relative position between aspect and context words. 

\paragraph{Setup.} We experimented on two datasets from SemEval 2014~\cite{absc}, which consist of reviews and comments from two categories: \textsc{Laptop} and \textsc{Restaurant}. We adopt the standard evaluation metrics: Accuracy and Macro-Averaged F1. We follow the instructions of \citet{zhang2019aspect} to run the experiments 5 times with random initialization and report the averaged performance. We denote the original PWCN with relative position information as PWCN-Pos, and that utilizes dependency trees constructed by SpaCy\footnote{https://spacy.io/} as PWCN-Dep. SpaCy has reported an UAS of 94.5 on English PTB and so it can serve as a good reference for human-designed dependency schema. We also compare our model against two trivial trees (left-chain and right-chain trees). For our model, we feed the corpus into BERT and extract dependency trees with the best performing setting: Eisner+Dist. For parsing, we introduce an inductive bias to favor short dependencies~\cite{eisner2010favor}. To ensure a fair comparison, we induce the root word from the impact matrix $\mathcal{F}$ instead of using the gold root. Specifically, we select the root word $x_k$ based on the simple heuristic $\argmax_i \sum_{j=1}^T f(x_i, x_j)$. 

\paragraph{Results.} Table~\ref{tab:sentiment} presents the performance of different models. We observe that the trees induced from BERT is either on-par (\textsc{Laptop}) or marginally better (\textsc{Restaurant}) in terms of downstream task's performance when comparing with trees produced by SpaCy. 
\textsc{Laptop} is considerably more difficult than \textsc{Restaurant} due to the fact that the sentences are generally longer, which makes inducing dependency trees more challenging.
We also see that the Eisner trees consistently perform better than the right-/left- chain baselines
, which leads to an exciting future work that investigates how encoding structural knowledge can help ABSC.

Our results suggest that although the tree structures BERT learns can disagree with parser-provided-linguistically-motivated ones to a large extent, they are also empirically useful to downstream tasks, at least to ABSC. As future work, we plan to extend our analysis to more downstream tasks and models, like those reported in Shi~\shortcite{shi2018tree}.

\begin{table}[]
    \centering
    \resizebox{\columnwidth}{!}{
    \begin{tabular}{lcccc}
    \toprule
    \multirow{2}{*}{\textbf{Model}}     & \multicolumn{2}{c}{\textbf{Laptop}} & \multicolumn{2}{c}{\textbf{Restaurant}}\\
                &  Acc       & Macro-F1      & Acc       & Macro-F1 \\
     \hline
    LSTM        & 69.63      & 63.51         & 77.99     & 66.91 \\ 
    \hline
    \textbf{PWCN} & & & &\\
    $\quad$+Pos    & 75.23      & 71.71        & 81.12      & 71.81 \\
    $\quad$+Dep    & 76.08      & 72.02        & 80.98      & 72.28 \\
    $\quad$+Eisner & 75.99      & 72.01        & \textbf{81.21}      & \textbf{73.00} \\
    $\quad$+right-/left-chain & 74.39  & 70.78        & 80.82      & 72.71 \\
    \toprule
    \end{tabular}
    }
    \caption{Experimental results of aspect based sentiment classification.}
    \label{tab:sentiment}
\end{table}
\section{Related Work}
There has been substantial research investigating what pre-trained language models have learned about languages' structures. 

One rising line of research uses probing classifiers to investigate the different syntactic properties captured by the model. They are generally referred to as ``probing task''~\cite{conneau2018you}, ``diagnostic classifier''~\cite{giulianelli2018under}, and ``auxiliary prediction tasks''~\cite{Adi2017}. The syntactic properties investigated range from basic ones like sentence length~\cite{Shi2016, Jawahar2019}, syntactic tree depth~\cite{Jawahar2019}, and segmentation~\cite{liu2019linguistic} to challenging ones like syntactic labeling~\cite{tenney2019bert,Tenney2019}, dependency parsing~\cite{Hewitt2019, Clark2019}, and constituency parsing~\cite{peters2018dissecting}. However, when a probe achieves high accuracy, it’s difficult to differentiate if it is the representation that encodes targeted syntactic information, or it is the probe that just learns the task~\cite{hewitt2019designing}.

In line with our work, recent studies seek to find correspondences between parts of the neural network and certain linguistic properties, without explicit supervision.

Most of them focus on analyzing attention mechanism, by extracting syntactic tree for each attention head and layer individually~\cite{Tiedemann2018, Clark2019}. Their goal is to check if the attention heads of a given pre-trained model can track syntactic relations better than chance or baselines. In particular, \citet{Tiedemann2018} analyze a machine translation model's encoder by extracting dependency trees from its self-attention weights, using Chu-Liu/Edmonds algorithm. \citet{Clark2019} conduct a similar investigation on BERT, but the simple head selection strategy they used does not guarantee a valid dependency tree. \citet{Marecek2018} propose heuristic methods to convert attention weights to syntactic trees. However, they do not quantitatively evaluate their approach. In their later study~\cite{Marecek2019}, they propose a bottom-up algorithm to extract constituent trees from transformer-based NMT encoders and evaluate their results on three languages. \citet{htut2019attention} reassess these works but find that there are no generalist heads that can do holistic parsing. Hence,  analyzing attention weights directly may not reveal much of the syntactic knowledge that a model has learned. Recent dispute about attention as explanation~\cite{jain2019attention, serrano-smith-2019-attention, wiegreffe2019attention} also suggests that the attention's behavior does not necessarily represent that of the original model. 

Another group of research examine the outputs of language models on carefully chosen input sentences~\cite{Goldberg2019, Bacon2019}. They extend previous works~\cite{Linzen2016, Gulordava2018, marvin2018targeted} on subject-verb agreement test (generating the correct number of a verb far away from its subject) to provide a measure of the model’s syntactic ability. Their results show that the BERT model captures syntax-sensitive agreement patterns well in general. However, subject-verb agreement cannot provide more nuanced tests of other complex structures (e.g., dependency structure, constituency structure), which are the interest of our work.

Two recent works also perturb the input sequence for model interpretability~\cite{Rosa2019, Xintong2019}. However, these works only perturb the sequence once. \citet{Rosa2019} utilize the original MLM objective to estimate each word's ``reducibility'' and import simple heuristics into a right-chain baseline to construct dependency trees. \citet{Xintong2019} focus on evaluating word alignment in NMT, but unlike our two-step masking strategy, they only replace the token of interest with a zero embedding or a randomly sampled word in the vocabulary.

\section{Discussion \& Conclusion} 
One concern shared by our reviewers is that performance of our probes are underwhelming: the induced trees are barely closer to linguist-defined trees than simple baselines (e.g., right\-branching) and are even worse in the case of discourse parsing. However, this does not mean that supervised probes are wrong or that BERT captures less syntax than we thought. In fact, there is actually no guarantee that our probe will find a strong correlation with human-designed syntax, since we do not introduce the human-designed syntax as supervision. What we found is the ``natural'' syntax inherent in BERT, which is acquired from self-supervised learning on plain text. We would rather say our probe complements the supervised probing findings in two ways. First, it provides a lower-bound (on the unsupervised syntactic parsing ability of BERT). By improving this lower-bound, we could uncover more ``accurate'' information to support supervised probes' findings. Second, we show that when combined with a down-stream application (sec~\ref{sec:sentiment}), the syntax learned by BERT might be empirically helpful despite not totally identical to the human design.

In summary, we propose a parameter-free probing technique to complement current line of work on interpreting BERT  through probes. With carefully designed two-stage perturbation, we obtain impact matrices from BERT. This matrix mirrors the function of attention mechanism that captures inter-word correlations, except that it emerges through the output of BERT model, instead of from intermediate representations. We devise algorithms to extract syntactic trees from this matrix. Our results reinforce those of~\cite{Hewitt2019, liu2019linguistic, Jawahar2019, Tenney2019, tenney2019bert} who demonstrated that BERT encodes rich syntactic properties. We also extend our method to probe document structure, which sheds lights on BERT's effectiveness in modeling long sequences. Finally, we find that feeding the empirically induced dependency structures into a downstream system~\cite{zhang2019aspect} can further improve its accuracy. The improvement is compatible with or even superior to a human-designed dependency schema. This offers an insight into BERT's success in downstream tasks. We leave it for future work to use our technique to test other linguistic properties (e.g., coreference) and to extend our study to more downstream tasks and systems.
\section{Acknowledgement}
We would like to thank Lingpeng Kong from DeepMind for his constructive feedback of the paper. This research is supported by Hong Kong Research Grant Council GRF grants 17254016.

\bibliography{acl2019}
\bibliographystyle{acl_natbib}

\end{document}